\documentclass[runningheads]{llncs}

\usepackage[T1]{fontenc}
\usepackage[utf8]{inputenc}
\usepackage{graphicx}
\usepackage{amsmath}
\usepackage{amssymb}
\usepackage{mathtools}
\usepackage{xifthen}
\usepackage{xfrac}
\usepackage{csquotes}
\usepackage{subcaption}
\usepackage[pdfborder={0 0 0}]{hyperref}
\usepackage{algorithm}
\usepackage[noend]{algpseudocode}
\usepackage{tikz}
\usetikzlibrary{patterns,intersections,calc}
\pgfdeclarelayer{bg}    %
\pgfsetlayers{bg,main}  %
\usepackage{todonotes}

\makeatletter
\newcommand{\tikzdistance}[3]{
\tikz@scan@one@point\pgfutil@firstofone($#1-#2$)\relax  
\pgfmathsetmacro{#3}{round(0.99626*veclen(\the\pgf@x,\the\pgf@y)/0.0283465)/1000}
}
\makeatother

\renewcommand{\refeq}[1]{(\ref{#1})}
\newcommand{\refsec}[1]{Sect.~\ref{#1}}
\newcommand{\reffig}[1]{Fig.~\ref{#1}}

\DeclareMathOperator*{\argmax}{arg\,max}

\newcommand{\eps}{\varepsilon}
\newcommand*\diff{\mathop{}\!\mathrm{d}}
\newcommand{\expect}[2][]{%
	\ifthenelse{\isempty{#1}}%
	{\operatornamewithlimits{\mathbb{E}}}%
	{\operatornamewithlimits{\mathbb{E}}_{#1}}%
	\left[#2\right]%
}
\newcommand{\norm}[1]{\left\Vert #1 \right\Vert}
\newcommand{\sqnorm}[1]{\norm{#1}^2}
\newcommand{\lowsuper}[1]{\textsuperscript{$#1$}}%
\newcommand{\mdot}[2]{\left\langle #1, #2 \right\rangle}
\newcommand{\trace}[1]{\operatorname{tr}\left(#1\right)}
\newcommand{\eye}[1][]{%
	\ifthenelse{\isempty{#1}}%
	{\mathrm{I}}%
	{\mathrm{I}_{#1}}%
}
\newcommand{\zeromat}[1][]{%
	\ifthenelse{\isempty{#1}}%
	{\mathbf{0}}%
	{\mathbf{0}_{#1,#1}}%
}
\newcommand{\zerovec}[1][]{%
	\ifthenelse{\isempty{#1}}%
	{\mathbf{0}}%
	{\mathbf{0}_{#1}}%
}
\newcommand{\half}[1][]{%
	\ifthenelse{\isempty{#1}}%
	{\sfrac{1\kern-1pt}{2}}%
	{\sfrac{#1}{2}}%
}

\usepackage{tikz}
\usetikzlibrary{calc}

\newcommand*{\highlightboxcolor}{gray}
\makeatletter
\newcommand{\highlightbox}[2][\highlightboxcolor]{%
	\!\textcolor{#1}{%
	\tikz[baseline={([yshift=-.5ex]current bounding box.center)}]%
	\node[%
		rectangle,%
		minimum width=1ex,%
		rounded corners,%
		fill=#1%
	] {\!\normalcolor\m@th$\displaystyle#2\vphantom{Tq}$\!};%
	}\!%
}
\makeatother

\begin{document}
\title{%
	On Projections to Linear Subspaces%
	\thanks{%
		Part of the work on this paper has been supported by Deutsche Forschungsgemeinschaft (DFG), project number 124020371, within the Collaborative Research Center SFB 876 ``Providing Information by Resource-Constrained Analysis'', project A2%
	}%
}%
\author{Erik Thordsen\inst{1}\orcidID{0000-0003-1639-3534} \and
Erich Schubert\inst{1}\orcidID{0000-0001-9143-4880}}
\authorrunning{E. Thordsen, E. Schubert}
\institute{TU Dortmund University, Otto-Hahn-Straße 14, 44227 Dortmund, Germany}
\maketitle
\begin{abstract}
The merit of projecting data onto linear subspaces is well known from, e.g., dimension reduction.
One key aspect of subspace projections, the maximum preservation of variance (principal component analysis), has been thoroughly researched and the effect of random linear projections on measures such as intrinsic dimensionality still is an ongoing effort.
In this paper, we investigate the less explored depths of linear projections onto explicit subspaces of varying dimensionality and the expectations of variance that ensue.
The result is a new family of bounds for Euclidean distances and inner products.
We showcase the quality of these bounds as well as investigate the intimate relation to intrinsic dimensionality estimation.

\end{abstract}

\section{Introduction}

\begin{tikzpicture}[overlay, remember picture]
\node[red, yshift=-20mm, anchor=north, text width=110mm] at (current page.north) {
\textbf{Preprint} version. Please consult the final version of record instead:
\\
Erik Thordsen, Erich Schubert:
On Projections to Linear Subspaces.\\
Similarity Search and Applications (SISAP 2022)\\
\url{https://doi.org/10.1007/978-3-031-17849-8_7}
};
\end{tikzpicture}%
The probably most important research on linear subspace projections was written by Pearson\nocite{pearson1901liii} in his 1901 paper on Principal Component Analysis (PCA). %
The concept of PCA explains how the variance of a data set can be decomposed into orthogonal components, each of which covers the maximum amount of variance. %
This fundamental result has been employed in many fields including dimensionality reduction, clustering~\cite{DBLP:conf/sdm/AchtertBKKZ07}, intrinsic dimensionality estimation~\cite{DBLP:journals/tc/FukunagaO71}, and many more.
The decomposition also implies linear projections that preserve the least amount of variance.
Yet, it yields little information on the less tangible middle ground of random projections.
The Johnson-Lindenstrauss lemma shows that random projections can preserve distances well, and 
the effect of random projections on, e.g., intrinsic dimensionality \cite{DBLP:conf/sisap/HouleK21} has also been explored in the past.
But we could not find literature on the effect of random projections on the variance itself.
In this paper, we investigate the effect on a projected point's squared norm which entails effects on the variance of the data set.
The arising bounds for the Euclidean distance as well as for inner products are explored in \refsec{sec:indexing}.
The projections required for these bounds rely on the normal vectors of the linear subspace on which we project, which are drawn from the data set itself.
Using measures based on points from the data set to assess boundaries on norms is a concept already employed in, e.g., spatial indexing.
Methods like LAESA \cite{DBLP:journals/prl/MicoOV94} use so-called pivot/reference/prototype points and the triangle inequality to prune the data set during spatial queries.
Tree-based methods like the Balltree \cite{omohundro1989five} use the triangle inequality to exclude entire subtrees,
while permutation based indexing~\cite{DBLP:journals/pami/ChavezFN08,DBLP:conf/sisap/VadicamoGA21} uses the relative closeness to reference points to partition the data.
The central points in these approaches fulfill a role equivalent to pivots.
Using pivots for random projections, however, yields fundamentally stronger pruning capabilities, as discussed in \refsec{sec:indexing}.
In \refsec{sec:spectral}, we analyze the expected values of variance preserved by random projections.
These expectations are closely related to PCA, yet costly to compute exactly.
To compensate for the computational cost and fathom the relation to eigenvalues we propose an approximation of the expected values in terms of eigenvalues.
The expected values are related to the Angle-Based Intrinsic Dimensionality (ABID) estimator~\cite{DBLP:conf/sisap/ThordsenS20}.
We explore the relationship in \refsec{sec:id_estimation}, which leads to a tangible link between indexing complexity and intrinsic dimensionality.
To highlight the practical implications as well as showcase the efficacy of the introduced bounds we propose a very simple index and our empirical results in \refsec{sec:evaluation}.
Lastly, we close with a summary of this paper and a short outlook on future research in \refsec{sec:conclusion}.

In this paper, we denote the $i$-th eigenvalue of some matrix $M$ with $\smash{\lambda^{(M)}_i}$.
We do not care about the specific order of eigenvalues but assume that corresponding eigenvalues of matrices that admit the same eigenvectors are in the same order.
We write $M^c$ as an abbreviation for $\smash{V\Lambda^cV^T}$ where $V$ is the matrix containing the eigenvectors of $M$ as columns and $\Lambda^c$ is the diagonal matrix containing $\smash{(\lambda_i^{(M)})^c}$ on the diagonal.
We write $C(X)$ for the covariance matrix of data sets $X$ where we assume $X$ to be origin-centered unless otherwise specified.
We denote the normalizations of vectors $x$ and data sets $X$ with $\widetilde{x}$ and~$\widetilde{X}$, respectively.
Whenever Euclidean spaces and distances are discussed, the dot product is implied by the inner product.

\section{Pivotal Bounds In Euclidean Spaces}
\label{sec:indexing}
We consider linear subspace projections of \textit{query points} onto the linear subspace spanned by (not necessarily orthogonal) \textit{pivots} or \textit{reference points}~$\{r_1, \ldots, r_k\}$, $k\,{\leq}\,d$ drawn from the same distribution as the analyzed data set,
e.g., by choosing them from the data set itself.
In the case of affine subspace projections, both the query and reference points are shifted by a \textit{center} point~$c$. %
We assume all (shifted) reference points to be linearly independent.
Otherwise, we discard reference points until linear independence holds.
The projection~$\pi(x{-}c;\,r_1{-}c,\ldots,r_k{-}c)$ of some shifted query point~$x{-}c$ onto the affine subspace (shortened to~$\pi(x{-}c)$ whenever the choice of reference points is clear) is then given by
\begin{align}
	\pi(x-c) &= \sum\nolimits_{i=1}^k \mdot{x-c}{\hat{r}_i} \hat{r}_i
	\label{eq:projection}
\end{align}
where the~$\hat{r}_i$ are the normalized orthogonal vectors obtained from the Gram-Schmidt process applied to the~$r_i{-}c$.
These can be recursively computed from
\begin{align}
	\hat{r}_1 &= \frac{r_1-c}{\norm{r_1-c}}&
	\hat{r}_i &= \frac
	{(r_i-c) - \sum\nolimits_{j=1}^{i-1} \mdot{r_i-c}{\hat{r}_j} \hat{r}_j}
	{\norm{(r_i-c) - \sum\nolimits_{j=1}^{i-1} \mdot{r_i-c}{\hat{r}_j} \hat{r}_j}}
	\label{eq:recursive_hat_ri}
\end{align}
where~$\norm{x}$ is shorthand for~$\smash{\mdot{x}{x}^{\half}}$.
In the following, we will repeatedly require the evaluation of~$\mdot{\cdot}{\hat{r}_i}$ and~$\norm{\pi(\cdot; \cdot)}$.
Although \refeq{eq:projection} and \refeq{eq:recursive_hat_ri} can be evaluated explicitly every time, it can be more convenient to represent the (squared) norm after projection in terms of inner products (especially in kernel spaces):
\begin{align}
	\sqnorm{\pi(x-c)} &= \sum\nolimits_{i=1}^k \mdot{x-c}{\hat{r}_i}^2
\end{align}
since all~$\hat{r}_i$ are normalized and pairwise orthogonal.
We can reduce~$\mdot{\cdot}{\hat{r}_i}$ to
\begin{align}
	\mdot{x-c}{\hat{r}_i}
	&= \tfrac{
		\mdot{c}{c} - \mdot{c}{x} - \mdot{c}{r_i} + \mdot{x}{r_i}
		- \sum\nolimits_{j=1}^{i-1} \mdot{x - c}{\hat{r}_j} \mdot{r_i - c}{\hat{r}_j}
	}{
		\left(
			\mdot{c}{c} - 2 \mdot{c}{r_i} + \mdot{r_i}{r_i}
			- \sum\nolimits_{j=1}^{i-1} \mdot{r_i - c}{\hat{r}_j}^2
		\right)^{\sfrac{1}{2}}
	}
	\label{eq:rhat_dot}\\
	\intertext{
		which can also be used recursively to compute the~$\mdot{r_i - c}{\hat{r}_j}$ in \refeq{eq:rhat_dot}.
		In the non-affine case,~$c = \zerovec$, \refeq{eq:rhat_dot} simplifies to
	}
	\mdot{x}{\hat{r}_i}
	&= \tfrac{
		\mdot{x}{r_i}
		- \sum\nolimits_{j=1}^{i-1} \mdot{x}{\hat{r}_j} \mdot{r_i}{\hat{r}_j}
	}{
		\left(
			\mdot{r_i}{r_i}
			- \sum\nolimits_{j=1}^{i-1} \mdot{r_i}{\hat{r}_j}^2
		\right)^{\sfrac{1}{2}}
	}
\end{align}
Note that the denominator and parts of the nominator need to be computed just once.
Further, we omit the explicit computation of any~$\hat{r}_i$ which would be infeasible in, e.g., RBF kernel and general inner product spaces.
With dynamic programming, $\sqnorm{\pi(x-c)}$ can be computed in~$\Theta(p k^2)$ time, where~$p$ is the effort required to compute an inner product.

In spatial indexing, pivots have been successfully used to bound distances via the triangle inequality \cite{DBLP:journals/prl/MicoOV94,omohundro1989five}.
We propose to bound distances in terms of a decomposition of the squared Euclidean norm into dot products given by
\begin{equation}
	d_{Euc}(x,y)^2 = \sqnorm{x-y} = \mdot{x-y}{x-y} = \mdot{x}{x} + \mdot{y}{y} - 2\mdot{x}{y}
	\label{eq:sqnorm_decomp}
\end{equation}
From this we can derive bounds for the Euclidean distance between two points given a bound on the dot product~$\mdot{x}{y}$, assuming~$\mdot{x}{x}$ and~$\mdot{y}{y}$ are known.
Let~$\hat{r}_1, \ldots, \hat{r}_k$ be pivot points previously orthogonalized by the Gram-Schmidt process as defined in \refsec{sec:spectral}.
We can decompose~$x-c$ and~$y-c$ into~$k$ components aligned along the~$\hat{r}_i$ and one orthogonal remainder.
We will call this~$(k+1)$-th component~$x_\bot$ and~$y_\bot$, respectively.
It then follows that
\begin{align}
	\mdot{x-c}{y-c}
	&= \mdot{x_\bot}{y_\bot}
	+ \sum\nolimits_{i=1}^k \mdot{\mdot{x-c}{\hat{r}_i}\hat{r}_i}{\mdot{y-c}{\hat{r}_i}\hat{r}_i}
\end{align}
Because the~$\hat{r}_i$ are pairwise orthogonal, this decomposition is uniquely defined.
Since all~$\hat{r}_i$ have a unit norm, we can rewrite this equation to
\begin{align}
	\mdot{x}{y}
	&= \mdot{x_\bot}{y_\bot}
	+ \mdot{c}{x} + \mdot{c}{y} - \mdot{c}{c}
	+ \sum\nolimits_{i=1}^k \mdot{x-c}{\hat{r}_i}\mdot{y-c}{\hat{r}_i}
	\label{eq:}
\end{align}
All of the terms on the right-hand side then either depend on~$x$ or~$y$, but not on both, except for~$\mdot{x_\bot}{y_\bot}$.
In the semantics of Euclidean spaces, both~$x_\bot$ and~$y_\bot$ lie in the same~$(d-k)$-dimensional linear subspace.
We can compute both as~${x_\bot=(x-c) - \pi(x-c)}$ and~${y_\bot=(y-c) - \pi(y-c)}$, respectively, but do not know their relative orientation.
Yet, we can bound their inner product using the Cauchy-Schwarz inequality resulting in the bounds~$\pm (\mdot{x_\bot}{x_\bot}\cdot\mdot{y_\bot}{y_\bot})^{\half}$.
By~orthogonality of~$x_\bot$ and~$\pi(x-c)$ we know~${\sqnorm{x_\bot}\,{=}\,\sqnorm{x-c} - \sqnorm{\pi(x-c)}}$.
The bounds for the inner product~$\mdot{x-c}{y-c}$ then follow as
\begin{align}
	& \begin{array}[t]{l}
		\mdot{c}{x} + \mdot{c}{y} - \mdot{c}{c}
		+ \sum\nolimits_{i=1}^k \mdot{x-c}{\hat{r}_i}\mdot{y-c}{\hat{r}_i}\\
		\pm \left(\begin{array}{l}
			\phantom{\cdot}\left(\mdot{x}{x} + \mdot{c}{c} - 2\mdot{c}{x} - \sum\nolimits_{i=1}^k \mdot{x-c}{\hat{r}_i}^2\right) \\
			\cdot\left(\mdot{y}{y} + \mdot{c}{c} - 2\mdot{c}{y} - \sum\nolimits_{i=1}^k \mdot{y-c}{\hat{r}_i}^2\right)
		\end{array}\right)^{\half}
	\end{array}
\end{align}
which in the non-affine case,~$c=\zerovec$, becomes
\begin{align}
	\sum\limits_{i=1}^k \mdot{x}{\hat{r}_i}\mdot{y}{\hat{r}_i}
	\pm \left(
		\left(\mdot{x}{x} - \sum\limits_{i=1}^k \mdot{x}{\hat{r}_i}^2\right) \cdot
		\left(\mdot{y}{y} - \sum\limits_{i=1}^k \mdot{y}{\hat{r}_i}^2\right)
	\right)^{\half}
	\label{eq:dot_bounds}
\end{align}
Inserting both of these values into \refeq{eq:sqnorm_decomp} gives bounds on the squared Euclidean distance and, consequentially, on the Euclidean distance.
These bounds are a generalization of at least two bounds known from the literature.
When we assume the affine case and~$k\,{=}\,0$ pivots, the bounds derived from \refeq{eq:sqnorm_decomp} and \refeq{eq:dot_bounds} reduce to
\begin{align}
	& \mdot{x}{x} + \mdot{y}{y} - 2\mdot{c}{x} - 2\mdot{c}{y} + 2\mdot{c}{c} \pm 2\norm{x-c}\norm{y-c}\\
	=& \left(\norm{x-c} \pm \norm{y-c}\right)^2
	\label{eq:triangle_ineq}
\end{align}
which are the bounds easily derivable from the triangle inequality.
For the non-affine case with $k\,{=}\,1$ pivots and normalized~$x$ and~$y$, the inner product bounds~\refeq{eq:dot_bounds} reduce to
\begin{align}
	\mdot{x}{\hat{r}_1} \mdot{y}{\hat{r}_1} \pm \left(
		\left(1 - \mdot{x}{\hat{r}_1}^2\right)
		\left(1 - \mdot{y}{\hat{r}_1}^2\right)
	\right)^{\half}
	\label{eq:cos_triangle_ineq}
\end{align}
which is the triangle inequality for cosines introduced in \cite{DBLP:conf/sisap/Schubert21}.
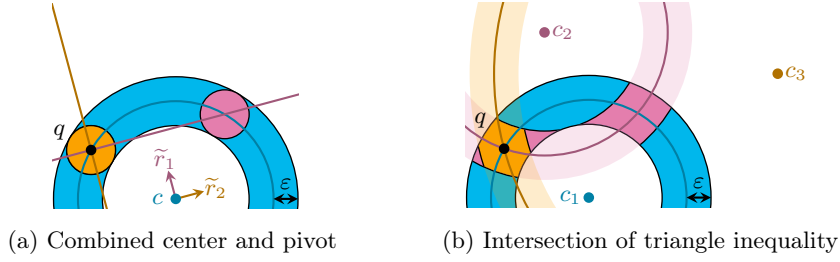
\begin{figure}[t]
	\centering
	\definecolor{ccolor}{HTML}{00B7EC}
	\definecolor{racolor}{HTML}{E47EAD}
	\definecolor{rbcolor}{HTML}{FBA200}
	\def\ccolor{ccolor!70!black}
	\def\racolor{racolor!70!black}
	\def\rbcolor{rbcolor!70!black}
	\def\fccolor{ccolor}
	\def\fracolor{racolor}
	\def\frbcolor{rbcolor}
	\def\qangle{150}
	\def\qradius{4em}
	\def\fradius{1em}
	\def\rlength{1.2em}
	\def\raangle{105}
	\def\epslen{.25pt}
	\ifthenelse{\raangle < 90}{
		\def\rbangle{\the\numexpr \raangle+90 \relax}
	}{
		\def\rbangle{\the\numexpr \raangle-90 \relax}
	}
	\def\clipping{
		\clip[shift=(C)] ({-\qradius-\fradius-\epslen},{-0.1*\qradius-0*\fradius}) rectangle ({\qradius+\fradius+\epslen},{1.75*\qradius+\fradius});
	}
	\begin{subfigure}[b]{.49\linewidth}\centering
		\begin{tikzpicture}[thick]
			\coordinate (C) at (0,0);
			\coordinate (Q) at ([shift=(\qangle:\qradius)] C);
			\coordinate (R1) at ([shift=(\raangle:\rlength)] C);
			\coordinate (R2) at ([shift=(\rbangle:\rlength)] C);
			\draw[\racolor, -stealth] (C) -- (R1);
			\draw[\rbcolor, -stealth] (C) -- (R2);
			\begin{scope}
				\clipping
				\draw[draw=\ccolor,thick,name path=CP] (C) circle (\qradius);
				\draw[\racolor,name path=RAP] ([shift=(\rbangle:-3*\qradius)] Q) -- ([shift=(\rbangle:3*\qradius)] Q);
				\draw[\rbcolor,name path=RBP] ([shift=(\raangle:-3*\qradius)] Q) -- ([shift=(\raangle:3*\qradius)] Q);
				\begin{pgfonlayer}{bg}
					\clipping
					\draw[draw=black,line width=2*\fradius+2*\epslen] (C) circle (\qradius);
					\draw[draw=\fccolor,line width=2*\fradius-2*\epslen] (C) circle (\qradius);
					\fill[name intersections={of=CP and RAP, total=\t}][black]
					\foreach \s in {1,...,\t} {(intersection-\s) circle (\fradius+\epslen)};
					\fill[name intersections={of=CP and RAP, total=\t}][\fracolor]
					\foreach \s in {1,...,\t} {(intersection-\s) circle (\fradius-\epslen)};
					\fill[black] (Q) circle (\fradius+\epslen);
					\fill[\frbcolor] (Q) circle (\fradius-\epslen);
				\end{pgfonlayer}
			\end{scope}
			\draw[black, stealth-stealth, shift=(C)] (\qradius,0) -- (\qradius+\fradius, 0);
			\fill[\ccolor] (C) circle (2pt);
			\fill[black] (Q) circle (2pt);
			\node[\ccolor, xshift=-.75em] at (C) {$c$};
			\node[shift=(\qangle:\fradius+.5em)] at (Q) {$q$};
			\node[\racolor, shift=(\raangle:.5em)] at (R1) {$\widetilde{r}_1$};
			\node[\rbcolor, shift=(\rbangle:.5em)] at (R2) {$\widetilde{r}_2$};
			\node[black] at ([shift=(C)] \qradius+.5*\fradius, .6em) {$\eps$};
		\end{tikzpicture}
		\caption{Combined center and pivot}
	\end{subfigure}
	\hfill
	\begin{subfigure}[b]{.49\linewidth}\centering
		\begin{tikzpicture}[thick]
			\coordinate (Q) at (0, 3em);
			\coordinate (R1) at (0, 0);
			\coordinate (R2) at ([shift=(\raangle:1.75*\qradius)] R1);
			\coordinate (R3) at ([shift=(\rbangle:2*\qradius)] Q);
			\coordinate (Q) at ([shift=(\qangle:\qradius)] R1);
			\tikzdistance{(R1)}{(Q)}{\raqd}
			\tikzdistance{(R2)}{(Q)}{\rbqd}
			\tikzdistance{(R3)}{(Q)}{\rcqd}
			\begin{scope}
				\clipping
				\draw[draw=black,line width=2*\fradius+2*\epslen] (R1) circle (\raqd);
				\draw[draw=\fccolor,line width=2*\fradius-2*\epslen] (R1) circle (\raqd);
				\draw[draw=\fracolor,opacity=.2,line width=2*\fradius] (R2) circle (\rbqd);
				\draw[draw=\frbcolor,opacity=.2,line width=2*\fradius] (R3) circle (\rcqd);
				\begin{scope}[even odd rule]
					\clip (R1) circle (\fradius+\raqd cm-\epslen) (R1) circle (-\fradius+\raqd cm+\epslen);
					\draw[draw=black,line width=2*\fradius+2*\epslen] (R2) circle (\rbqd cm);
					\draw[draw=\fracolor,line width=2*\fradius-2*\epslen] (R2) circle (\rbqd cm);
				\end{scope}
				\begin{scope}[even odd rule]
					\clip (R1) circle (\fradius+\raqd cm-\epslen) (R1) circle (-\fradius+\raqd cm+\epslen);
					\clip (R2) circle (\fradius+\rbqd cm-\epslen) (R2) circle (-\fradius+\rbqd cm+\epslen);
					\draw[draw=black,line width=2*\fradius+2*\epslen] (R3) circle (\rcqd cm);
					\draw[draw=\frbcolor,line width=2*\fradius-2*\epslen] (R3) circle (\rcqd cm);
				\end{scope}
				\draw[draw=\ccolor,thick,name path=RAP] (R1) circle (\raqd);
				\draw[draw=\racolor,thick,name path=RBP] (R2) circle (\rbqd);
				\draw[draw=\rbcolor,thick,name path=RCP] (R3) circle (\rcqd);
			\end{scope}
			\draw[black, stealth-stealth, shift=(C)] (\qradius,0) -- (\qradius+\fradius, 0);
			\fill[\ccolor] (R1) circle (2pt);
			\fill[\racolor] (R2) circle (2pt);
			\fill[\rbcolor] (R3) circle (2pt);
			\fill[black] (Q) circle (2pt);
			\node[shift=(\qangle-20:\fradius+.5em)] at (Q) {$q$};
			\node[\ccolor, shift=(180:.75em)] at (R1) {$c_1$};
			\node[\racolor, shift=(0:.75em)] at (R2) {$c_2$};
			\node[\rbcolor, shift=(0:.75em)] at (R3) {$c_3$};
			\node[black] at ([shift=(C)] \qradius+.5*\fradius, .6em) {$\eps$};
		\end{tikzpicture}
		\caption{Intersection of triangle inequality}
	\end{subfigure}
	\caption{
		Eligible search spaces around a query point $q$ after filtering with the lower bounds obtained from \highlightbox[ccolor!60!white]{one}, \highlightbox[racolor!60!white]{two}, or \highlightbox[rbcolor!60!white]{three} centers and/or pivots.
	}
	\label{fig:volume_comparison}
\end{figure}
Triangle-inequality-based bounds have been used in spatial indexing in methods like, e.g., LAESA~\cite{DBLP:journals/prl/MicoOV94}.
For multiple pivots, these approaches take the minimum or maximum of the bounds obtained separately for each pivot.
In our terminology, we refer to such pivots as centers~$c$.
Those are fundamentally different from the term pivots introduced here:
When performing an~$\eps$-range query for a query point~$y$, the eligible search space for vectors~$x$ according to the upper bound in \refeq{eq:triangle_ineq} is a hyperspherical shell centered at~$c$.
This geometric shape can be described as the sumset (the set of all sums of pairs in the cartesian product) of a~$(d{-}1)$-sphere of radius~$\norm{y{-}c}$ centered at~$c$ and a~$d$-ball of radius~$\eps$.
When using pivots as per our definition, each pivot induces a hyperplane orthogonal to the~$\hat{r}_i$ which intersects with the hypersphere. %
Consequentially, the resulting eligible search space is the sumset of a~$(d{-}1{-}k)$-sphere of radius~$(\sqnorm{y{-}c} {-} \sqnorm{\pi(y{-}c)})^{\half}$ and a~$d$-ball of radius~$\eps$.
This is illustrated in two dimensions in \reffig{fig:volume_comparison}.
Each of the pivots eliminates an entire dimension from the sphere-part of the search space whereas the minimum lower bounds obtained from multiple centers produce an intersection of multiple hyperspherical shells.
While~$d{-}1$ pivots can reduce the search space to the sumset of at most 2 points and an~$\eps$-ball, the intersection of even~$d$ hyperspherical shells in the best case produces a volume that can be roughly described as a distorted hypercube with an \enquote{edge length} of about~$2\eps$.
The resulting volume can be exponentially larger in~$d$ than the search volume using~$d{-}1$ pivots.
As the volumes of regular shapes in Euclidean space expand exponentially in dimensions, one would expect an approximately exponential reduction in search space over an increasing number of pivots, whereas using the minimum upper bound over multiple centers does not induce such a reduction in search space volume.
It is, therefore, of little surprise that the cosine bounds introduced in~\cite{DBLP:conf/sisap/Schubert21} ($k{=}1$), produced tighter bounds empirically than the triangle inequality ($k{=}0$), and were successfully applied to improve the performance of spherical k-means clustering~\cite{DBLP:conf/sisap/SchubertLF21}.
Qualitatively, there is a clear argument for using a larger amount of pivots.
However, the reduction in search space comes at the price of increased computational cost as the evaluation of~$\mdot{y}{\hat{r}_i}$ is quadratic and the evaluation of the bounds is linear in~$k$.
Blindly increasing~$k$ is not universally advantageous for the computational cost of spatial indexing queries.
But how many pivots tighten the bounds enough to counterweigh the overhead?
More precisely, how much more of a point's squared norm does the~$k$-th randomly drawn pivot drawn cover on average?
Although the answer does not refer to an optimal pivot choice, by arguing over expectations of underlying distributions, this conservative argument likely holds for previously unknown query points.

\section{Expected Variance Of Random Projections}
\label{sec:spectral}

The analysis of squared norms after projection is closely related to spectral analysis.
If we chose any normalized vector~$v$, $\expect[x \in X]{\smash{\sqnorm{\pi(x-\expect[y \in X]{y}; v)}}}$ is simply the variance of $X$ in direction $v$.
Consequentially, for any pair of a normalized eigenvector $e_i$ and the corresponding eigenvalue $\smash{\lambda_i\lowsuper{(C(X))}}$, we know that $\expect[x \in X]{\smash{\sqnorm{\pi(x; e_i)}}} {=} \smash{\lambda_i\lowsuper{(C(X))}}$ for any origin-centered $X$.
By orthogonality of the eigenvectors, this argument can be extended to any number of eigenvectors $e_1, \ldots, e_n$ as
\begin{equation}
	\expect[x \in X]{\sqnorm{\pi(x; e_1, \ldots, e_n)}} = \sum\nolimits_{i=1}^n \lambda_i^{(C(X))}
\end{equation}
Pearson \cite{pearson1901liii} showed that the eigenvectors of the covariance matrix are precisely the maximizers of this term, i.e. they are the solution to
\begin{eqnarray}
	\argmax_{e_1, \ldots, e_n}~\expect[x \in X]{\sqnorm{\pi(x; e_1, \ldots, e_n)}}
\end{eqnarray}
If one intended to evaluate how much of the squared norm of any point is remaining after the projection onto $k$ directions maximally, the answer immediately follows from the sum of the $k$ largest eigenvalues.
Employing the corresponding eigenvectors as $\hat{r}_i$ would then be a reasonable approach.
Yet, both eigenvectors and eigenvalues can be sensitive to noise in limited data sets \cite{DBLP:journals/tsp/EversonR00}.
They may not be an optimal choice when new and unknown data arises.
We, hence, focus on the expectation of these values for a random set of reference points drawn from the data.
More precisely we inspect
\begin{eqnarray}
	E^\Sigma_k(X) &:=&
	\expect[\substack{r_1, \ldots, r_k \in X\\\forall i \neq j: r_i \neq r_j}]{
		\expect[x \in X]{\sqnorm{\pi(x-c; r_1-c, \ldots, r_n-c)}}
	}
	\label{eq:explained_squared_norm}
\end{eqnarray}
As with the eigenvectors and eigenvalues of the covariance matrix, this expected value is the sum of components introduced by each additional reference point taken into consideration.
This naturally sums up the total variance of the data set for $k=d$.
Through varying $k$ we can obtain a cumulative description of how much variance an arbitrary linear projection within the data set can explain and the difference of neighboring values gives the amount of variance explained at random by the $k$-th component.
We will write this difference as $E_k(X) := E^\Sigma_k(X) - E^\Sigma_{k-1}(X)$ where $E^\Sigma_0(X) = 0$.
It follows that $E^\Sigma_k(X) = \sum_{i=1}^k E_k(X)$.
Practically evaluating the expected value from any data set $X$ for any $k \gg 1$ is infeasible, as it involves $\binom{\vert X \vert}{k}$ possible sets of reference points.
It is much easier to estimate the value by the Monte Carlo method (i.e. choosing a fixed number of random sets of reference points) or to approximate it from the covariance matrix if it well describes the data set's distribution.

We will only consider the non-affine case of $c=\zerovec$, as the affine case is analogous and introduces numerous subtractions hindering readability.
We will also omit the constraint that the reference points must not be linearly dependent to improve readability.
Starting from \refeq{eq:explained_squared_norm} we can deduce
\begin{align}
	E_k(X) &= E^\Sigma_k(X) - E^\Sigma_{k-1}(X)%
	= \expect[\substack{x \in X,\\r_1, \ldots, r_k \in \widetilde{X}}]{
		\mdot{x}{\tfrac{
			r_k - \pi(r_k; r_1, \ldots, r_{k-1})
		}{
			\norm{r_k - \pi(r_k; r_1, \ldots, r_{k-1})}
		}}^2
	}\\
	\intertext{Here the term $r_k - \pi(r_k; r_1, \ldots, r_{k-1})$ is the projection of $r_k$ onto the linear subspace orthogonal to all $r_1, \ldots, r_{k-1}$.
	We can represent this projection by a matrix multiplication with a matrix, which we will call $A_{k-1}$.}
	&= \expect[\substack{x \in X,\\r_1, \ldots, r_k \in \widetilde{X}}]{
		\tfrac{\mdot{x}{A_{k-1} r_k}^2}
		{\mdot{A_{k-1} r_k}{A_{k-1} r_k}}
	}
	= \expect[\substack{x \in X,\\r_1, \ldots, r_k \in \widetilde{X}}]{
		x^T\tfrac{A_{k-1} r_k r_k^T A_{k-1}^T}
		{\trace{A_{k-1} r_k r_k^T A_{k-1}^T}}
		x
	}\\
	\intertext{By rewriting $r_i r_i^T$ as $R_i$ this further simplifies to}
	&= \expect[\substack{x \in X,\\r_1, \ldots, r_k \in \widetilde{X}}]{
		x^T\tfrac{A_{k-1} R_k A_{k-1}^T}
		{\trace{A_{k-1} R_k A_{k-1}^T}}
		x
	}
	\\
	&= \trace{
		\expect[r_1, \ldots, r_{k-1} \in \widetilde{X}]{
			\expect[r_k \in X]{
				\tfrac{A_{k-1} R_k A_{k-1}^T}
				{\trace{A_{k-1} R_k A_{k-1}^T}}
			}
		}
		\expect[x \in X]{
			xx^T
		}
	}\\
	\intertext{By replacing $\expect[x \in X]{xx^T}$ with the covariance matrix $C(X)$ and renaming the innermost expected value to $C_k(X)$ we then obtain}
	&= \expect[r_1, \ldots, r_{k-1} \in \widetilde{X}]{\trace{C_k(X) C(X)}}
\end{align}
$A_0$ is the identity matrix $\eye[d]$, as the linear subspace orthogonal to an empty set of vectors is the entire space.
Consequentially, we can define $A_k$ recursively as
\begin{eqnarray}
	A_k
	&=& A_{k-1} - \tfrac{A_{k-1}R_{k}A_{k-1}^T}{\trace{A_{k-1}R_{k}A_{k-1}^T}}
	= A_{k-1} - \tfrac{A_{k-1}R_{k}A_{k-1}}{\trace{A_{k-1}R_{k}A_{k-1}}}
\end{eqnarray}
As all $R_i$ are symmetric, all $A_i$ are symmetric as well. %
The expected value over $r_k$ of $\tfrac{A_{k-1} R_k A_{k-1}}{\trace{A_{k-1}R_k A_{k-1}}}$ now (approximately) equals the covariance matrix of $X$ after being projected to the linear subspace orthogonal to $r_1, \ldots, r_{k-1}$ and normalized.
It follows immediately that $C_1(X) {=} C(\widetilde{X})$ and thereby $E_1(X) = \trace{\smash{C(\widetilde{X})C(X)}}$.
However, $E_k(X)$ for $k>1$ is much less easily defined because the $A_i$ are dependent on the effective values of all $r_j$, $j \leq i$, and not only on~$r_i$.
To circumvent the problem we assume that all $A_i$ are aggregate matrices just like $C(X)$ and sufficiently independent of each other to evaluate the $C_k(X)$ recursively.
To highlight this assumption we will denote the approximated $A_i$ as a function of $X$ as $A_i(X)$.
We further assume that all $A_i(X)$, $C_i(X)$, and $C(X)$ admit the same eigenvectors, whereby
\begin{align}
	E_k(X)
	&= \expect[r_1, \ldots, r_{k-1} \in \widetilde{X}]{\trace{C_k(X)C(X)}}
	= \sum\nolimits_{i=1}^d \lambda_i^{(C_k(X))} \lambda_i^{(C(X))}
	\label{eq:Ek_approx}
\end{align}
We will hereafter omit the $(X)$ in superscripts of eigenvalues for readability.
Although the resulting values are no longer exact due to these two assumptions, they allow us to approximate the expected value by deriving the value of~$\lambda_i\lowsuper{(C_k)}$.
Assuming that $X$ is multivariate normally distributed, we can extract this value from the definition of~$C_k(X)$ using the corresponding eigenvector~$e_i$:
\begin{align}
	\lambda_i^{(C_k)}
	&= e_i^T C_k(X) e_i = \trace{e_ie_i^TC_k(X)}\\
	&= \expect[r_k \in X]{
		\tfrac{r_k^T A_{k-1}(X) e_ie_i^T A_{k-1}(X) r_k}
		{r_k^T A_{k-1}(X)^2 r_k}
	}\\
	&= \expect[r_k \in \mathcal{N}_{\zerovec[d],\eye[d]}]{
		\tfrac{r_k^T C(X)^{\half} A_{k-1}(X) e_ie_i^T A_{k-1}(X) C(X)^{\half} r_k}
		{r_k^T C(X)^{\half} A_{k-1}(X)^2 C(X)^{\half} r_k}
	}\\
	&= \expect[r_k \in \mathcal{N}_{\zerovec[d],\eye[d]}]{
		\tfrac{r_k^T e_ie_i^T C(X) A_{k-1}(X)^2 r_k}
		{r_k^T C(X) A_{k-1}(X)^2 r_k}
	}\\
	\intertext{We now substitute $C(X) A_{k-1}(X)^2$ with $D_{k-1}(X)$ which entails $\smash{\lambda_j\lowsuper{(C)}\left(\lambda_j\lowsuper{(A_{k-1})}\right)^2}$ is equal to $\lambda_j\lowsuper{(D_{k-1})}$.
	In favor of brevity we will omit the exponent $(D_{k-1})$ from here on.
	As per Proposition 2 in Kan and Bao \cite{DBLP:journals/ma/BaoK13}, $\lambda_i\lowsuper{(C_k)}$ then equals}
	&= \int_0^\infty \frac{
		\trace{e_ie_i^T D_{k-1}(X) (\eye[d]+2t D_{k-1}(X))^{-1}}
	}{
		\vert \eye[d] + 2t D_{k-1}(X) \vert^{\half}
	} \diff t\\
	&= \int_0^\infty \frac{
		\lambda_i
	}{
		\left(1 + 2t \lambda_i \right)^{\half}
		\prod_{j=1}^d \left(1 + 2t \lambda_j \right)^{\half}
	} \diff t
\end{align}
This integral is closely related to elliptic integrals and we do not provide a simple and closed-form solution.
Solving the integral numerically would again involve too much computational effort.
We instead propose to substitute the $\lambda_j$ in the denominator with $(\lambda_i^2 \prod_{j=1}^d \lambda_j)^{\sfrac{1}{(d{+}2)}}$ whereby the integral takes the form of a scaled beta prime distribution:
\begin{align}
	\lambda_i^{(C_k)}
	\approx &~\lambda_i B(\alpha,\beta) \int_0^\infty \tfrac{t^{\alpha-1}\big(1+2\left(\lambda_i^2 \prod_{j=1}^d \lambda_j\right)^{\frac{1}{d+2}}t\big)^{-\alpha-\beta}}{B(\alpha,\beta)} \diff t\\
	\intertext{where $\alpha = 1$, $\beta = \frac{d}{2}$, and $B(\alpha,\beta)$ is the beta function.
	The integral over the scaled beta distribution is known to equal the scaling factor, whereby}
	\lambda_i^{(C_k)}
	\approx &~\tfrac{\lambda_i B(\alpha,\beta)}{2\left(\lambda_i^2 \prod_{j=1}^d \lambda_j\right)^{\frac{1}{d+2}}}
	\quad \propto \lambda_i^{\frac{d}{d+2}}
	\label{eq:approxNormed}
\end{align}
As the $\smash{\lambda_i^{(C_k)}}$ are eigenvalues of a normalized distribution, their sum must equal~1.
Using this constraint, we can drop all factors independent of $\lambda_i$ and derive
\begin{align}
	\lambda_i^{(C_k)}
	&\approx \lambda_i^{\frac{d}{d+2}} \Big/ \sum\nolimits_{j=1}^d \lambda_j^{\frac{d}{d+2}}
\end{align}
As the $\lambda_j$ are dependent on $\smash{\lambda_j^{(C)}}$ and $\smash{\lambda_j^{(A_{k-1})}}$, this leads to the recursive definition
\begin{align}
	\lambda_i^{(C_k)}
	&\approx \tfrac{\big(\lambda_i\lowsuper{(C)}\big(\lambda_i\lowsuper{(A_{k-1})}\big)^2\big)^{\frac{d}{d+2}}}{\sum_{j=1}^d \big(\lambda_j\lowsuper{(C)}\big(\lambda_j\lowsuper{(A_{k-1})}\big)^2\big)^{\frac{d}{d+2}}}&
	\lambda_i^{(A_k)}
	&\approx \lambda_i^{(A_{k-1})} - \lambda_i^{(C_{k-1})}
	\label{eq:approx}
\end{align}
This recursion terminates at $\lambda_i^{(A_0)} = 1$ and $\lambda_i^{(C_0)} = 0$.
These approximations can be computed efficiently in $\Theta(dk)$ and inserted in \refeq{eq:Ek_approx} to give an approximation of~$E_k(X)$.
Since the approximations are based on the assumption that $X$ is distributed according to some multivariate normal distribution they need not be accurate.
Since all occurrences of any $r_k$ in the formulae involve some sort of normalization, this approximation extends to any distribution of $X$ for which $\{C(X)^{-\half}x \mid x \in X\}$ is spherically symmetrically distributed, which includes cases like, e.g., $d$-balls.
We also did not compensate for the requirement that all $r_k$ must be pairwise different, as these arguments are based on distributions rather than point sets.
In empirical tests the sample size, however, did not contribute to approximation quality. %
The biggest issue with this approximation is the fact, that while the $A_i$ as variables in $r_1$ through $r_i$ must have eigenvalues in $\{0,1\}$, the approximated eigenvalues $\smash{\lambda_i\lowsuper{(A_k)}}$ can become negative whereby latter $E_k$ can be vastly overestimated.
As we know that the $E_k^\Sigma(X)$ must sum to the total variance of $X$, we propose to cut off any excess in $E_k^\Sigma(X)$ and determine the $E_k(X)$ based on these cut values.
To summarize, the approximation proceeds as follows:
For all $1 {\leq} k {\leq} d$ compute the $\smash{\lambda_i\lowsuper{(C_k)}}$ values using the recursive formulations~\refeq{eq:approx}.
Use these values to compute $E_i(X)$ values and reduce $E_i(X)$ values for larger $k$ to not have their sum exceed the total variance of $X$, which compensates for negative $\smash{\lambda_i\lowsuper{(A_k)}}$.
Even though this approximation from a theoretical point makes the wrong assumptions that the $r_k$ are pairwise different and that the $C_i(X)$ are statistically independent, the approximation in our experiments gave close enough results to have it worth considering, especially as the exact computation of values has an enormous computational cost.
The approximation via the Monte Carlo method is known to converge on the exact values, yet, might require enormous samples.%

While \refeq{eq:Ek_approx} requires the covariance matrix of a mean-centered data set, the approach via Monte Carlo sampling applies directly to inner product values and, hence, to kernel spaces.
The approximation in \refeq{eq:Ek_approx} can then be used in black-box optimization to obtain an approximate spectral analysis of the kernel space.
The obtained spectrum is neglecting the scale of the eigenvalues of the covariance matrix as the~$E_i(X)$ are invariant under the scaling of these values.
In this manner, we can perform approximate spectral analysis even in spaces that do not allow for a direct approach, such as the RBF kernel space which has infinitely many dimensions.
Naturally, the method must be applied in a truncated fashion for infinite dimensions, for which we here propose two solutions:
Firstly, one can estimate~$E_1(X)$ through~$E_k(X)$ for some fixed~$k$ using the Monte Carlo method and rescale these values to sum to 1.
This implies neglecting the remaining~$d{-}k$ dimensions and assuming the data to have 0 variance along with these directions.
The~$d{-}k$ smallest eigenvalues of the covariance of such a data set must then be 0, too.
Finding any set of~$k$ eigenvalues that leads to these~$E_1(X)$ through~$E_k(X)$ values then solves the truncated case.
Secondly, one can assume that the remaining variance not explained by~$E_k^\Sigma(X)$ is distributed over the remaining~$d{-}k$ values according to some user-defined distribution.
Assuming a uniform distribution, for example, would explain the remaining variance as noise in the embedding space which might be a reasonable assumption.

A special case can further be made on the evaluation of~$E_k(X)$ values on normalized data.
When working on~$\smash{\widetilde{X}}$ instead of~$X$, which can be achieved in kernel space by dividing the occurrences of~$x$ in the formulae by~$\mdot{x}{x}\!\lowsuper{\half}$, we immediately obtain that~$E_1(\smash{\widetilde{X}})$ equals the sum of squared eigenvalues of~$C(\smash{\widetilde{X}})$.
While this equality does not hold for the approximation via eigenvalues of~$C(\smash{\widetilde{X}})$, it is approximately obtained from the Monte Carlo method or precisely for an exhaustive evaluation of~$E_1(\smash{\widetilde{X}})$.
Just as the constraint of the sum of eigenvalues of~$C(\smash{\widetilde{X}})$ equalling~1, this additional constraint can be used in the black-box optimization for retrieving the original eigenvalues from~$E_k(\smash{\widetilde{X}})$ values.
Using~\refeq{eq:approxNormed}, these eigenvalues can be approximately translated into the relative eigenvalues of the non-normalized data whenever the data can be assumed to obey the distributional constraints of the approximation.

\section{Random Projections and ID Estimation}
\label{sec:id_estimation}

As stated in the previous section, $E_1(\smash{\widetilde{X}})$ equals the sum of squared eigenvalues of $C(\smash{\widetilde{X}})$.
The reciprocal of this specific value has been introduced as an estimator for intrinsic dimensionality named ABID \cite{DBLP:conf/sisap/ThordsenS20}, that is
\begin{equation}
	\operatorname{ID}_{\textit{ABID}}(X) = E_1(\widetilde{X})^{-1} = E_1^\Sigma(\widetilde{X})^{-1}
\end{equation}
For one, this observation adds additional semantics to the meaning of ABID as the number of basis vectors of a random projection to fully explain the variance in a data set.
Yet, it also implies the applicability of the $E_k$ values in the realm of ID estimation. %
Although $E_1$ gives the part of total variance a random projection based on in-distribution basis vectors can explain, not all $E_k$ values are necessarily equal.
That is, the projection onto two random directions does not necessarily cover twice the variance covered by projecting onto one random direction.
This linearity is exclusively true for spherically symmetrical distributions such as $d$-balls and for all other distributions we would certainly expect $E_2^\Sigma(X)~{<}~2 E_1^\Sigma(X)$.
Ultimately, we are looking for the smallest $k$ such that $E_k^\Sigma(X)~{\geq}~\trace{C(X)}$, that is, the number of random projections required to explain the entire variance of $X$.
Unfortunately, we only have formulae for integer $k$ but we can generalize the approach of ABID in the sense of extrapolating from a fixed $E_k$ which results in a parameterized ID estimator which we name the Thresholded Random In-distribution Projections (TRIP) Estimator:
\begin{equation}
	\operatorname{ID}_{\textit{TRIP}}(X,k,\eta) = k + \frac{(1-\eta) \trace{C(X)} - E_k^\Sigma(X)}{E_k(X)}
\end{equation}
where $k$ is the number of considered projections and $\eta \in [0,1]$ is a fraction describing how much of the variance we attribute to noise.
Semantically this answers the question \textit{\enquote{How many random projections are required to explain $(1{-}\eta)$ of the total variance if every further projection covers as much variance as the last one?}}.
In the linear case of spherically symmetrical distributions as above, this estimator is ideally constant for $\eta~{=}~0$ and all $1~{\leq}~k~{\leq}~d$.
On other distributions with $\eta~{=}~0$, we would expect a curve that starts at (approximately, dependent on implementation) $\operatorname{ID}_{\textit{ABID}}(X)$ for $k~{=}~1$ and approaches $k$ for increasing $k$ as the $E_i(X)$ are monotonically falling.
Equality is likely only reached for $k~{=}~d$, as this requires zero variance after $k$~projections, which is unlikely in presence of high-dimensional noise.
The factor~$\eta$ is intended to compensate for this.
For $\eta~{>}~0$, the curve again starts at approximately $\operatorname{ID}_{\textit{ABID}}(X)$, approaches $k$, and after some $k$ drops below it.
As for parameter choice, $\eta$ is application dependent whereas $k$ can either be chosen empirically, or we can inspect values $1~{\leq}~k~{\leq}~d$ to find the $k$ at which $\operatorname{ID}_{\textit{TRIP}}(X,k,\eta)$ is closest to $k$.
The latter is likely not feasible in a local ID fashion when using the Monte Carlo or exhaustive methods but can be done when using the approximation introduced in \refsec{sec:spectral}.
When using a fixed $k$, obtaining an ID below this $k$ is a strong indicator of having chosen $k$ too large.
In addition, the curve of $\operatorname{ID}_{\textit{TRIP}}(X,k,\eta)$ over varying~$k$, just like the curve of~$E_i(X)$, gives insights into the local distribution characteristics of the data set that goes beyond ID estimation.
These curves can theoretically help distinguish different subspaces, even when they share similar local ID.%

Referring back to the discussions of indexing with linear projections in \refsec{sec:indexing}, we can now state a clear connection between indexing with random in-distribution pivots and intrinsic dimensionality measures.
The $E_k^\Sigma(X)$ values answer how much variance on average is covered by a set of $k$ random pivots.
The expected covered variance is -- in an idealized case of, e.g., uniformly distributed hyperballs -- reciprocally related to intrinsic dimensionality.
This is most explicitly stated in the relation to ABID and gives rise to the TRIP estimator above.
Using this geometric concept of ID estimation, we can argue on an on-average appropriate number of pivots in spatial indexing.
In \refsec{sec:indexing} we observed that the eligible search space for range queries when using $k$ pivots is the sumset of a $(d-1-k)$-sphere and an $\eps$-ball.
The radius of the hypersphere is equal to the norm of the component orthogonal to all pivots,
and roughly describes how close the bounds derived in \refsec{sec:indexing} are to the true distances.
But there is a clear limit as to how much precision one needs in a finite data set.
If this radius drops below the distance between nearest points, removing this slack from the distance estimates does not improve the discriminability.
By choosing $\eta = \delta^2/\trace{C(X)}$ where $\delta$ is the, e.g., mean/median/$p$-percentile of nearest neighbor distances, we can use the TRIP estimator to evaluate just how many random projections exhaust the discriminative potential of pivoted indexing on average.

\section{Pivot Filtering Linear Scan}
\label{sec:evaluation}

\begin{figure}[t]
	\centering
	\begin{subfigure}[b]{.3\linewidth}
		\centering
		\includegraphics[width=\textwidth]{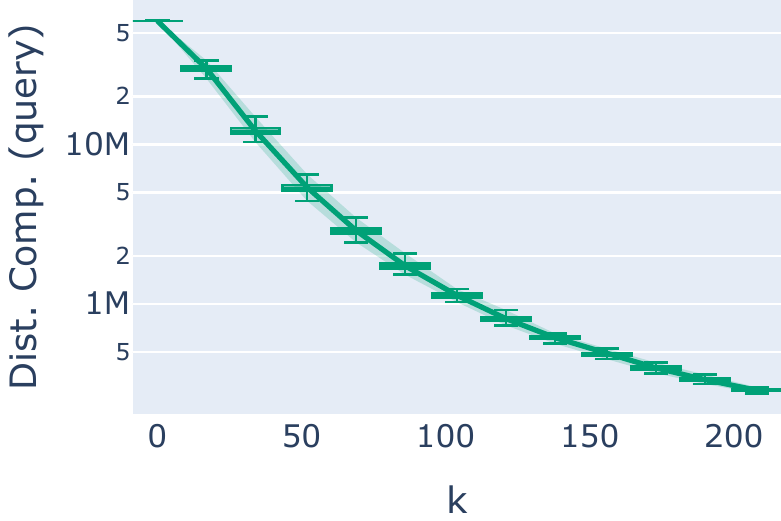}
		\vphantom{
			\hspace*{-.25\linewidth}
			\includegraphics[width=1.5\linewidth]{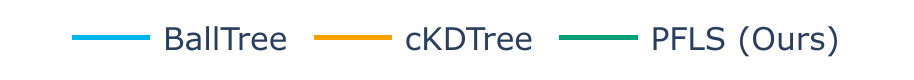}
		}
		\caption{Distance comp.}
		\label{fig:index_eval:dists}
	\end{subfigure}
	\begin{subfigure}[b]{.3\linewidth}
		\centering
		\includegraphics[width=\textwidth]{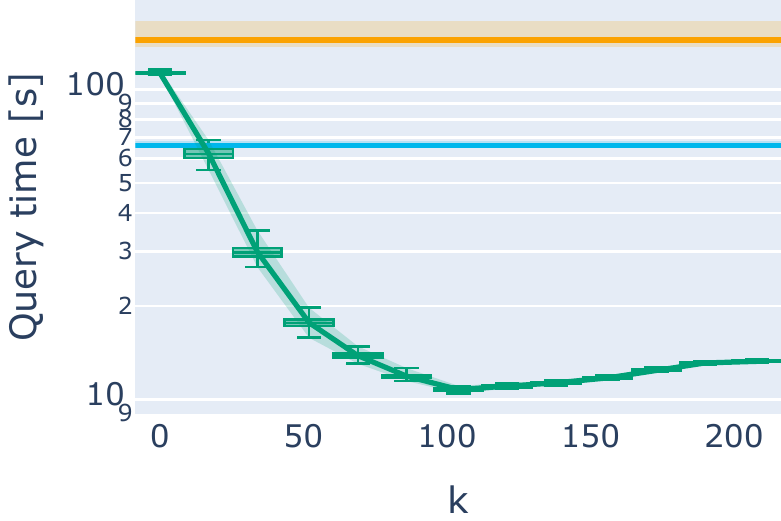}
		\hspace*{-.85\linewidth}
		\includegraphics[width=1.5\linewidth]{assets/plots/mnist/query_time_legend.pdf}
		\caption{Computation times}
		\label{fig:index_eval:time}
	\end{subfigure}
	\begin{subfigure}[b]{.325\linewidth}
		\centering
		\includegraphics[width=\textwidth]{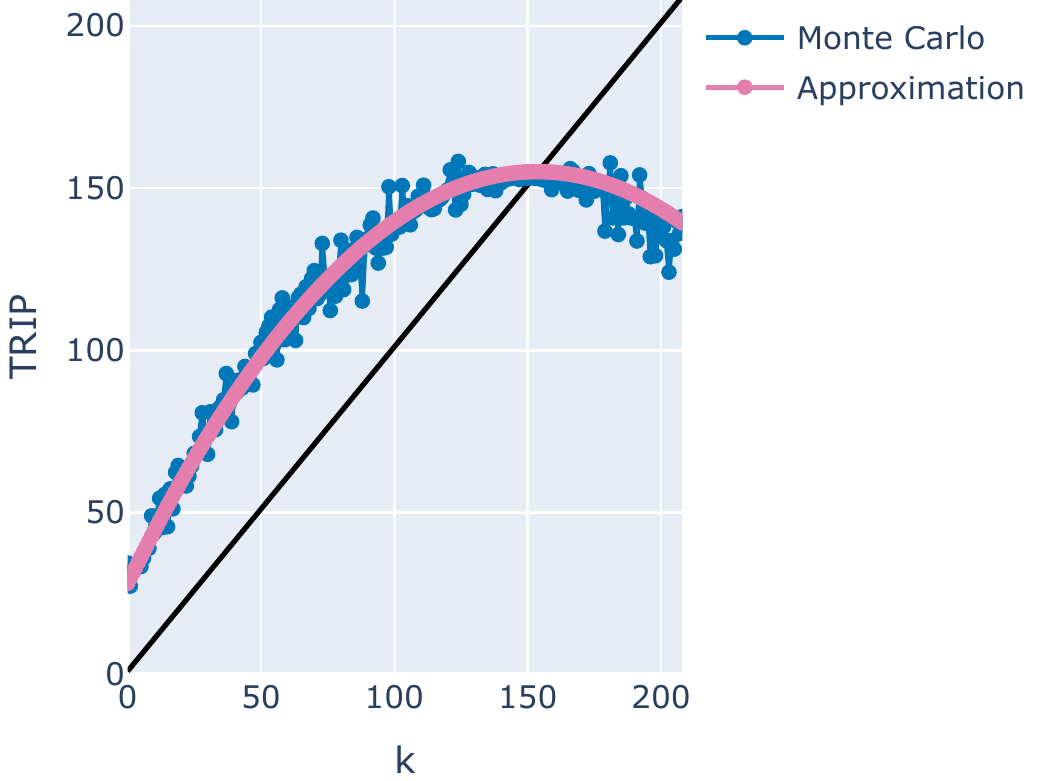}
		\caption{$\operatorname{ID}_{\textit{TRIP}}$ with $\eta > 0$}
		\label{fig:index_eval:trip}
	\end{subfigure}
	\caption{
		Experimental results on varying numbers of pivots.
		Additional pivots exponentially reduce the distance computations, but the query time stagnates once the average discriminative power of the bounds has been exploited.
		A suitable number of pivots is suggested at the crossing point of~$\operatorname{ID}_{\textit{TRIP}}$ with the diagonal.
		Lines are average values, shaded area indicates the minimum and maximum.
	}
	\label{fig:index_eval}
\end{figure}

For quality evaluation of the bounds as well as to validate the theoretical claims, we embed the bounds in a simple and easy-to-implement index.
During the initialization, we choose~$k$ random pivots. %
As mentioned in \refsec{sec:indexing}, we pre-compute all parts of the equations that are independent of query points such as~$\mdot{x}{\hat{r}_i}$ or the denominators in \refeq{eq:rhat_dot}.
Range and~$n$-nearest neighbor queries were then implemented according to Algorithms~\ref{alg:knn} and \ref{alg:range}.
The algorithms are quite similar to LAESA \cite{DBLP:journals/prl/MicoOV94} but do not require aggregation of multiple bounds as discussed in \refsec{sec:indexing}.
Both algorithms are at least linear in~$\vert X \vert$, which should be accounted for when comparing the performance with tree-based indices.
Integrating the bounds into a tree-based index is a nearby extension but out of the scope of this paper.
Both~Algorithms~\ref{alg:knn} and~\ref{alg:range} are trivially adaptable to search for the largest instead of the smallest distances.
This index is also trivially adaptable to work on inner products instead of distances by exchanging the bounds.
For our experiments, we implemented the index in the Rust language and called the functions from a Python wrapper to compare them to the cKDTree and BallTree implementations of SciPy~\cite{2020SciPy-NMeth}.
The source code is publicly available at \href{https://github.com/eth42/pfls}{\texttt{https://github.com/eth42/pfls}}.
\begin{algorithm}[bt]%
	\begin{algorithmic}%
		\algnewcommand{\IIf}[1]{\State\algorithmicif\ #1\ \algorithmicthen\ }
		\algnewcommand{\EndIIf}{\unskip}
		\algnewcommand{\IElsIf}[1]{\State\algorithmicelse\ \algorithmicif\ #1\ \algorithmicthen\ }
		\Function{query}{$y \in \mathbb{R}^d, n \geq 1$}%
		\State $ls \leftarrow$ lower bounds of $d(x,y)$ for all $x \in X$ as per \refeq{eq:sqnorm_decomp} and \refeq{eq:dot_bounds}%
		\State $h \leftarrow$ empty max heap%
		\State sort $X$ by ascending $ls[x]$
		\For{$x \in X$}%
			\If{$\vert h \vert < n$ \textbf{or} ($ls[x] < h.max.key$ \textbf{and} $d(x,y) < h.max.key$)}%
				\State push $x$ onto $h$ with key $d(x,y)$%
				\IIf{$\vert h \vert > n$}%
					remove entry with largest key from $h$%
				\EndIIf%
			\IElsIf{$ls[x] \geq h.max.key$}%
				\textbf{break}%
			\EndIf%
		\EndFor%
		\State \Return $h$ as array/list%
		\EndFunction%
	\end{algorithmic}%
	\caption{$n$-nearest neighbor query for distances}%
	\label{alg:knn}%
\end{algorithm}%
\begin{algorithm}[bt]%
	\begin{algorithmic}%
		\algnewcommand{\IIf}[1]{\State\algorithmicif\ #1\ \algorithmicthen\ }
		\algnewcommand{\EndIIf}{\unskip}
		\Function{query-range}{$y \in \mathbb{R}^d, \eps \in \mathbb{R}$}%
		\State $ls, hs \leftarrow$ lower and upper bounds of $d(x,y)$ for all $x \in X$ as per \refeq{eq:sqnorm_decomp} and \refeq{eq:dot_bounds}%
		\State $v \leftarrow$ empty list%
		\For{$x \in X$}%
			\IIf{$ls[x] < \eps$ \textbf{and} ($hs[x] < \eps$ \textbf{or} $d(x,y) < \eps$)}%
				Push $x$ into $v$%
			\EndIIf%
		\EndFor%
		\State \Return $v$%
		\EndFunction%
	\end{algorithmic}%
	\caption{range query for distances}%
	\label{alg:range}%
\end{algorithm}
Using this very simple index we investigated the theoretical claims and the quality of the bounds.
\reffig{fig:index_eval} displays the results of applying the index to the MNIST training data set.
All queries were 100-nearest-neighbor queries for 1000 query points drawn from the same data set.
We performed~100 queries for each set of parameters and instantiated a new index for each query.
As seen in \reffig{fig:index_eval:dists}, the number of distance computations initially drops exponentially as we increase the number of pivots,
which supports the theoretical claim that each pivot effectively eliminates one dimension from the data set and reduces the remaining search space exponentially.
For increasing~$k$, the descent in distance computations diminishes as the bounds become tight enough to sufficiently discriminate on neighboring points, and
the query time eventually increases due to the cost of computing the bounds.
In \refsec{sec:id_estimation}, we argued that the bounds only need to be as tight as to differentiate between nearest neighbors.
To validate this claim, we investigated the~$\operatorname{ID}_{\textit{TRIP}}$ values using an~$\eta$ equal to the 10-percentile of squared 1-nearest-neighbor distances divided by the total variance of the distribution.
The smallest~$k$ for which~$\operatorname{ID}_{\textit{TRIP}}(X, k, \eta) \leq k$ is around~150 as can be seen in \reffig{fig:index_eval:trip}.
The minimum computation time in \reffig{fig:index_eval:time} is around~100 but the query time at~$k~{=}~150$ is not that much larger than at~$k~{=}~100$.
The exact percentile is an educated guess and could be supported by inspecting the histogram of nearest-neighbor distances.
Yet, the region of~$k$ that provides low query times is wide enough that rough estimates and educated guesses are likely to give good results.
We conclude that~$\operatorname{ID}_{\textit{TRIP}}$ can be used to estimate a proper value for~$k$ by deriving~$\eta$ from a percentile of {1-nearest} neighbor distances.
To estimate a proper~$k$ efficiently, the approximation introduced in \refsec{sec:spectral} can be used, which practically is sufficiently similar to the values obtained from Monte Carlo sampling as displayed in \reffig{fig:index_eval:trip}.
\begin{figure}[t]
	\def\plotheight{7.5em}
	\def\legendwidth{1.5\linewidth}
	\def\legendvmargin{-.4em}
	\def\legend{
		\hspace*{-.25\linewidth}%
		\vspace*{\legendvmargin}%
		\includegraphics[width=\legendwidth]{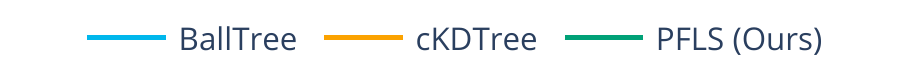}%
		\vspace*{\legendvmargin}%
	}
	\def\legendphantom{%
		\vspace*{\legendvmargin}%
		\vphantom{\legend}%
		\vspace*{\legendvmargin}%
	}
	\centering
	\begin{subfigure}[t]{.325\linewidth}
		\centering
		\includegraphics[height=\plotheight]{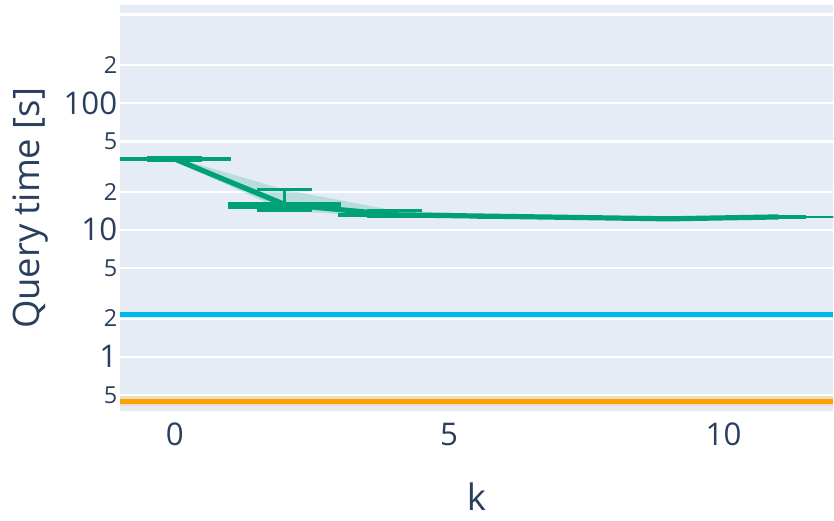}
		\legendphantom
		\caption{$3 \times 3 \times 3$ dim.}
		\label{fig:performance:aloi3}
	\end{subfigure}
	\begin{subfigure}[t]{.3\linewidth}
		\centering
		\includegraphics[height=\plotheight]{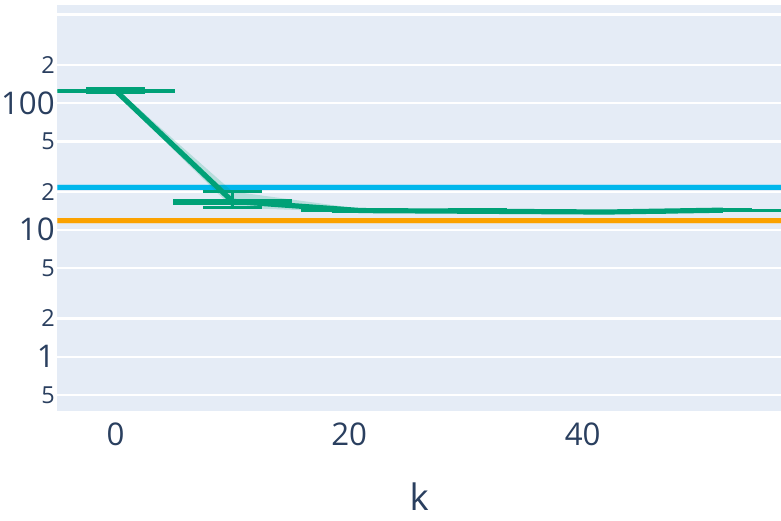}
		\legend
		\caption{$14 \times 3 \times 3$ dim.}
		\label{fig:performance:aloi14}
	\end{subfigure}
	\begin{subfigure}[t]{.3\linewidth}
		\centering
		\includegraphics[height=\plotheight]{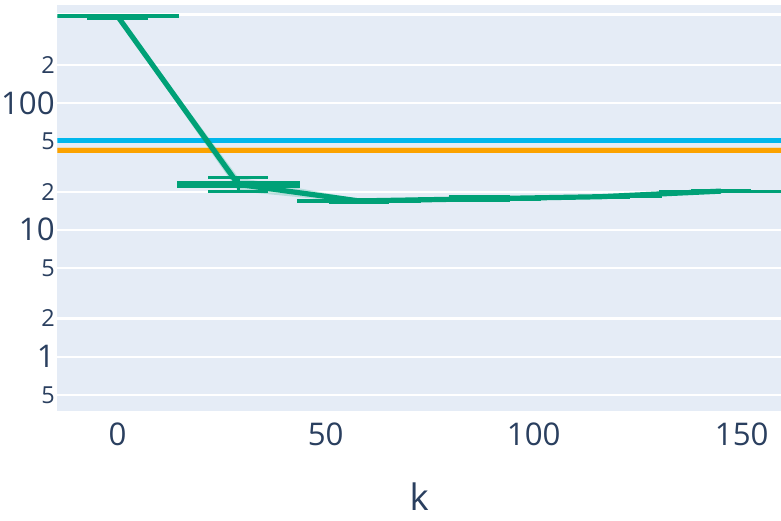}
		\legendphantom
		\caption{$14 \times 5 \times 5$ dim.}
		\label{fig:performance:aloi14x5}
	\end{subfigure}
	\caption{
		Query times for ALOI color histograms with varying dimensionality.
	}
	\label{fig:performance}
\end{figure}
Lastly, we compared query times on HSV color histograms of the ALOI data set with varying numbers of dimensions~\cite{aloi_data}.
The considered variants consist of~110250 instances with 27, 126, and 350 dimensions, respectively.
As can be seen in \reffig{fig:performance} the query performance of our index %
is mostly unaffected by increasing dimensionality.
Due to our index using a linear scan, the tree-based reference implementations were faster on low dimensionality.
For sufficiently high dimensional or small enough data sets, our index can outperform these reference implementations.
For larger data sets, extending the approach to a tree-based structure appears promising.

\section{Conclusion}
\label{sec:conclusion}

In this paper, we introduced new bounds for Euclidean distances and inner products using a pivot-based approach.
We showed that these bounds generalize the well-known bounds based on the triangle inequality.
We argued why an increased number of pivots exponentially reduces the eligible search space of certain queries and derived an approach to estimate a reasonable number of pivots for practical purposes.
We further showed how this number of pivots is intimately related to intrinsic dimensionality estimation.
Lastly, we implemented the bounds in a simple and easily reproducible index that operates on both inner products and their induced distances and allows queries for the smallest and largest values.
The empirical data presented aligns with the theoretical considerations and highlights the qualitative performance of implementing the bounds.
Further research should be invested in integrating these bounds into more sophisticated indices or constructing a tree-based index using these bounds.

\bibliographystyle{splncs04}
\bibliography{literature}

\begin{thebibliography}{10}
\providecommand{\url}[1]{\texttt{#1}}
\providecommand{\urlprefix}{URL }
\providecommand{\doi}[1]{https://doi.org/#1}

\bibitem{DBLP:conf/sdm/AchtertBKKZ07}
Achtert, E., B{\"{o}}hm, C., Kriegel, H., Kr{\"{o}}ger, P., Zimek, A.: Robust,
  complete, and efficient correlation clustering. In: {SIAM} Int. Conf. Data
  Mining ({SDM}). pp. 413--418 (2007). \doi{10.1137/1.9781611972771.37}

\bibitem{DBLP:journals/ma/BaoK13}
Bao, Y., Kan, R.: On the moments of ratios of quadratic forms in normal random
  variables. J. Multivar. Anal.  \textbf{117},  229--245 (2013).
  \doi{10.1016/j.jmva.2013.03.002}

\bibitem{DBLP:journals/pami/ChavezFN08}
Ch{\'{a}}vez, E., Figueroa, K., Navarro, G.: Effective proximity retrieval by
  ordering permutations. {IEEE} Trans. Pattern Anal. Mach. Intell.
  \textbf{30}(9),  1647--1658 (2008). \doi{10.1109/TPAMI.2007.70815}

\bibitem{DBLP:journals/tsp/EversonR00}
Everson, R.M., Roberts, S.J.: Inferring the eigenvalues of covariance matrices
  from limited, noisy data. {IEEE} Trans. Signal Process.  \textbf{48}(7),
  2083--2091 (2000). \doi{10.1109/78.847792}

\bibitem{DBLP:journals/tc/FukunagaO71}
Fukunaga, K., Olsen, D.R.: An algorithm for finding intrinsic dimensionality of
  data. {IEEE} Trans. Computers  \textbf{20}(2),  176--183 (1971).
  \doi{10.1109/T-C.1971.223208}

\bibitem{DBLP:conf/sisap/HouleK21}
Houle, M.E., Kawarabayashi, K.: The effect of random projection on local
  intrinsic dimensionality. In: Int. Conf. Similarity Search and Applications
  ({SISAP}). pp. 201--214 (2021). \doi{10.1007/978-3-030-89657-7\_16}

\bibitem{DBLP:journals/prl/MicoOV94}
Mic{\'{o}}, L., Oncina, J., Vidal, E.: A new version of the nearest-neighbour
  approximating and eliminating search algorithm {(AESA)} with linear
  preprocessing time and memory requirements. Pattern Recognit. Lett.
  \textbf{15}(1),  9--17 (1994). \doi{10.1016/0167-8655(94)90095-7}

\bibitem{omohundro1989five}
Omohundro, S.M.: Five balltree construction algorithms. International Computer
  Science Institute Berkeley (1989)

\bibitem{pearson1901liii}
Pearson, K.: On lines and planes of closest fit to systems of points in space.
  The London, Edinburgh, and Dublin philosophical magazine and journal of
  science  \textbf{2}(11),  559--572 (1901)

\bibitem{DBLP:conf/sisap/Schubert21}
Schubert, E.: A triangle inequality for cosine similarity. In: Int. Conf.
  Similarity Search and Applications ({SISAP}). pp. 32--44 (2021).
  \doi{10.1007/978-3-030-89657-7\_3}

\bibitem{DBLP:conf/sisap/SchubertLF21}
Schubert, E., Lang, A., Feher, G.: Accelerating spherical k-means. In: Int.
  Conf. Similarity Search and Applications ({SISAP}). pp. 217--231 (2021).
  \doi{10.1007/978-3-030-89657-7\_17}

\bibitem{aloi_data}
Schubert, E., Zimek, A.: {ELKI} multi-view clustering data sets based on the
  {A}msterdam library of object images ({ALOI}). Zenodo (2010).
  \doi{10.5281/zenodo.6355684}

\bibitem{DBLP:conf/sisap/ThordsenS20}
Thordsen, E., Schubert, E.: {ABID:} angle based intrinsic dimensionality. In:
  Int. Conf. Similarity Search and Applications ({SISAP}). pp. 218--232 (2020).
  \doi{10.1007/978-3-030-60936-8\_17}

\bibitem{DBLP:conf/sisap/VadicamoGA21}
Vadicamo, L., Gennaro, C., Amato, G.: On generalizing permutation-based
  representations for approximate search. In: Int. Conf. Similarity Search and
  Applications, {SISAP} (2021). \doi{10.1007/978-3-030-89657-7\_6}

\bibitem{2020SciPy-NMeth}
Virtanen, P., et~al.: {{SciPy} 1.0: Fundamental Algorithms for Scientific
  Computing in Python}. Nature Methods  \textbf{17},  261--272 (2020).
  \doi{10.1038/s41592-019-0686-2}

\end{thebibliography}

\end{document}